\documentclass{article} 
\usepackage{colm2024_conference}
\usepackage[misc]{ifsym}
\usepackage{microtype}
\usepackage{hyperref}
\usepackage{url}
\usepackage{booktabs}
\definecolor{darkblue}{rgb}{0, 0, 0.5}
\hypersetup{colorlinks=true, citecolor=darkblue, linkcolor=darkblue, urlcolor=darkblue}
\usepackage{multirow}
\usepackage{booktabs}
\usepackage{multicol}
\usepackage{appendix}
\usepackage{graphicx}
\usepackage{algorithm}
\usepackage{algorithmic}
\usepackage{amsmath}
\usepackage{amssymb}
\usepackage{amsfonts}
\usepackage{amsthm}
\definecolor{authorcolor}{RGB}{18, 101, 251}

\title{ImagebindDC: Compressing Multi-modal Data with Imagebind-based Condensation}

\author{
    \small
  {\bf
    Yue Min $^*$$^{\spadesuit,\clubsuit,\diamondsuit}$
    Shaobo Wang $^*$$^{\spadesuit}$
    Jiaze Li $^{\spadesuit}$ 
    Tianle Niu $^{\spadesuit}$
    Junxin Fan $^{\spadesuit}$    
    Yongliang Miao $^{\spadesuit}$
    \vspace{3pt}
  } \\
    \small
  {
  \bf
    Lijin Yang $^{\text{\Letter}}$$^{\clubsuit}$
    Linfeng Zhang
    $^{\text{\Letter}}$$^{\spadesuit}$
    \vspace{2pt}
  } \\
    \small
    {
    $^{\spadesuit}$ EPIC Lab, SJTU $\quad$
    $^{\clubsuit}$ Bosch Corporate Research Asia Pacific $\quad$
    $^{\diamondsuit}$ HKUST $\quad$
        \vspace{2pt}
  } \\
    \small{
  $^*$ Equal contribution $\quad$ 
    $^{\text{\Letter}}$ Corresponding authors
    }
}

\colmfinalcopy 
\begin{document}

\maketitle

\begin{abstract}
Data condensation techniques aim to synthesize a compact dataset from a larger one to enable efficient model training, yet while successful in unimodal settings, they often fail in multimodal scenarios where preserving intricate inter-modal dependencies is crucial. To address this, we introduce ImageBindDC, a novel data condensation framework operating within the unified feature space of ImageBind. Our approach moves beyond conventional distribution-matching by employing a powerful Characteristic Function (CF) loss, which operates in the Fourier domain to facilitate a more precise statistical alignment via exact infinite moment matching. We design our objective to enforce three critical levels of distributional consistency: (i) uni-modal alignment, which matches the statistical properties of synthetic and real data within each modality; (ii) cross-modal alignment, which preserves pairwise semantics by matching the distributions of hybrid real-synthetic data pairs; and (iii) joint-modal alignment, which captures the complete multivariate data structure by aligning the joint distribution of real data pairs with their synthetic counterparts. Extensive experiments highlight the effectiveness of ImageBindDC: on the NYU-v2 dataset, a model trained on just 5 condensed datapoints per class achieves lossless performance comparable to one trained on the full dataset, achieving a new state-of-the-art with an 8.2\% absolute improvement over the previous best method and more than 4$\times$ less condensation time.
\end{abstract}

\section{Introduction}

The remarkable success of modern AI has been largely propelled by the synergy between large-scale models and vast datasets~\citep{gpt3, vit, llama3, qwen3}. However, this paradigm comes at a significant cost: the computational, storage, and financial burdens associated with training on massive datasets are becoming prohibitive. Dataset Condensation (DC) has emerged as a compelling solution to this challenge~\citep{DD, DD_review}. The goal of DC is to synthesize a small, synthetic dataset that, while a fraction of the original size, preserves its essential information, enabling models to be trained to high performance with dramatically reduced resources. 

\begin{figure}[tb!]
    \centering
    \includegraphics[width=0.99\linewidth]{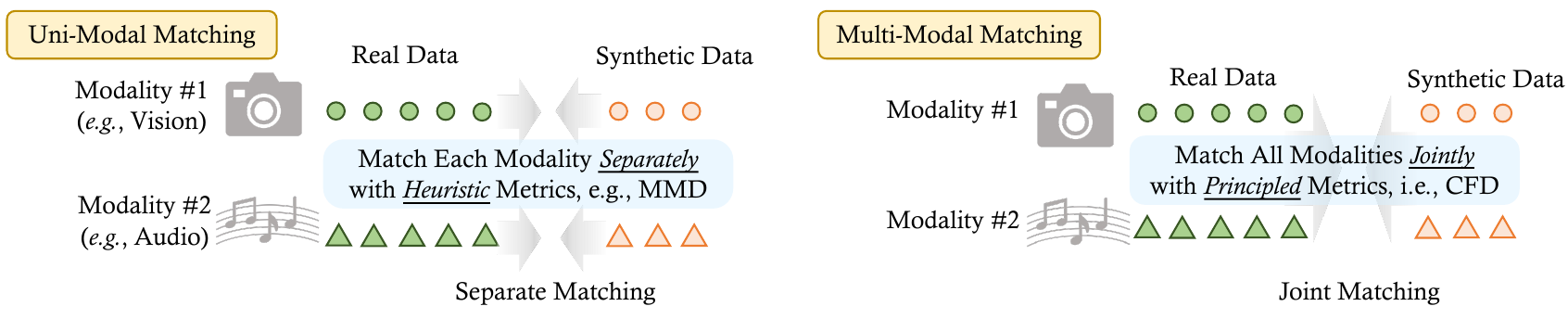}
    \caption{A Comparison of multi-modal Data Condensation Paradigms. 
    \textbf{(Top) Separate Matching:} Conventional methods condense each modality (\emph{e.g.}, vision, audio) independently, often using heuristic metrics like MMD. This preserves uni-modal statistics but critically fails to capture the cross-modal relationships that link the data together.
    \textbf{(Bottom) Joint Matching:} Our proposed framework, ImageBindDC, performs joint matching of all modalities simultaneously within a unified feature space. By using a principled metric like Characteristic Function Distance (CFD), our approach preserves the complete multi-modal data structure, ensuring the synthesized data is semantically coherent.}
    \label{fig:paradigm}
\end{figure}

Early successes in this domain have demonstrated the feasibility of condensing thousands of images into just a handful, accelerating training by orders of magnitude~\citep{DC, DSA, MTT, DATM, NCFM}. While effective in uni-modal settings (\emph{e.g.}, images-only), these traditional DC methods falter in the face of the increasingly prevalent multi-modal world. Modern applications frequently leverage rich, interconnected data from various sources, such as image, audio, text, and depth~\citep{CLIP, Imagebind}. The key challenge in condensing such data is not merely to preserve the statistical properties within each modality independently, but to maintain the intricate \textit{cross-modal relationships} that encode semantic meaning. As shown in Figure~\ref{fig:paradigm}, existing condensation techniques, designed with a uni-modal perspective, are ill-equipped for this task; they may create a statistically representative set of sythetic data, usually little datapoints per class (DPC), but they break the vital link that pairs a specific image with its corresponding sound.

\begin{figure}[tb!]
    \centering
    \includegraphics[width=0.99\linewidth]{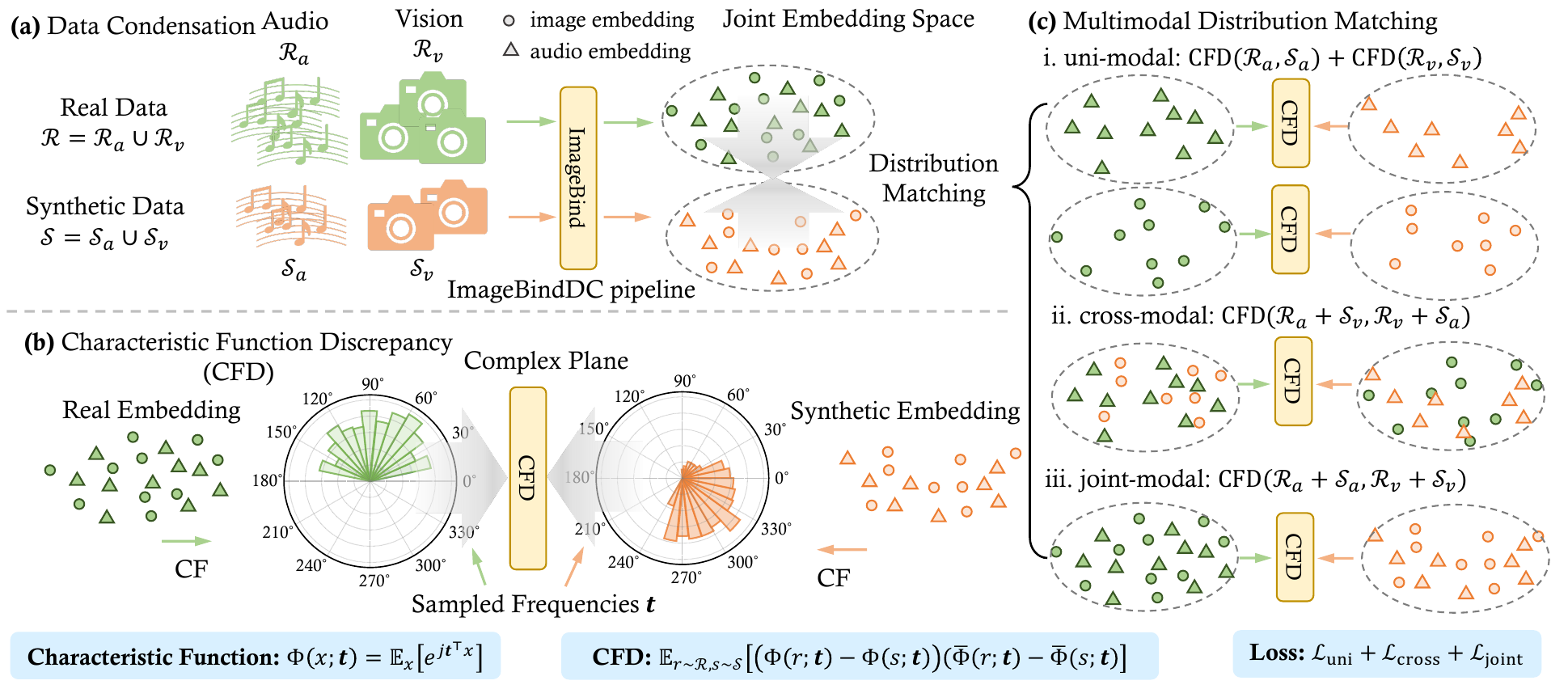}
    \caption{\textbf{Overview of the ImageBindDC Framework.} Our method condenses multi-modal data by performing principled distribution matching in a unified embedding space.
    \textbf{(a) Data Condensation Pipeline:} We take real multi-modal data, consisting of vision ($\mathcal{R}_v$) and audio ($\mathcal{R}_a$), and aim to synthesize a much smaller synthetic dataset ($\mathcal{S}_v, \mathcal{S}_a$). Both real and synthetic data are projected into a joint embedding space using the pretrained ImageBind encoder. The core of our method is to optimize the synthetic data such that its distribution in this embedding space matches that of the real data.
    \textbf{(b) Characteristic Function Discrepancy (CFD):} We use CFD as our distribution matching metric. The empirical Characteristic Function (CF) of a data distribution is calculated, which provides a summary in the Fourier domain (visualized here on the complex plane via polar plots). CFD then measures the discrepancy between the CFs of the real and synthetic embeddings, effectively matching all statistical moments for a precise alignment.
    \textbf{(c) Multi-modal Distribution Matching Objective:} To ensure comprehensive alignment, our final loss is a sum of three CFD-based objectives: 
    (\textbf{i}) \textit{Uni-modal alignment} preserves the integrity of each modality by matching real and synthetic data within the same modality (\emph{e.g.}, $\mathcal{R}_v$ vs. $\mathcal{S}_v$). 
    (\textbf{ii}) \textit{Cross-modal alignment} preserves the semantic relationship between modalities by matching the distribution of hybrid pairs (\emph{e.g.}, real audio + synthetic vision vs. real vision + synthetic audio).
    (\textbf{iii}) \textit{Joint-modal alignment} captures the complete data structure by matching the joint distribution of paired real data against paired synthetic data. The total loss $\mathcal{L}_{\text{total}} = \mathcal{L}_{\text{uni}} + \mathcal{L}_{\text{cross}} + \mathcal{L}_{\text{joint}}$ guides the synthesis process.
    }
    \label{fig:ImageBindDC}
\end{figure}

To bridge this critical gap, we introduce \textbf{ImageBindDC}, a novel framework for multi-modal data condensation. Our core insight is to perform condensation not in the raw, disparate data spaces, but within a \textbf{unified, joint-embedding space} provided by a large-scale pretrained model like ImageBind~\citep{Imagebind}. This allows us to directly address inter-modal dependencies. Furthermore, we move beyond conventional distribution matching techniques by employing a powerful \textbf{Characteristic Function (CF) loss}~\citep{NCFM}. Operating in the Fourier domain, our CF-based objective enables a more precise statistical alignment by matching an infinite number of moments between the synthetic and real data distributions. As illustrated in Figure~\ref{fig:ImageBindDC}, our approach enforces distributional consistency at three critical levels: uni-modal, cross-modal, and joint-modal, ensuring that the synthesized data captures the complete multi-modal structure. Our main contributions are summarized as follows:
\begin{itemize}
    \item We propose \textbf{ImageBindDC}, the first data condensation framework specifically designed to operate in a unified feature space, effectively preserving complex multi-modal data relationships.
    \item We design a novel, multi-faceted objective that leverages a Characteristic Function (CF) based loss to enforce three critical levels of statistical consistency: \textbf{uni-modal}, \textbf{cross-modal}, and \textbf{joint-modal alignment}, which together preserve the complete multi-modal data structure.
    \item Extensive experiments demonstrate the state-of-the-art performance of ImageBindDC, achieving the best results across various multimodal datasets. Notably, a model trained on just 20 synthesized image-audio pairs reached 98\% of the full-dataset performance in Audio-Visual Event Localization classification, marking a 2.53\% improvement. Also, on NYU-v2 Dataset, ImageBindDC achieved 8.2\% absolute improvement over the method second in performance. Our method also shows exeptional computational efficiency, reducing dataset condension time by over 4.6$\times$ for 20 DPC.
\end{itemize}

\section{Related Work}
\label{sec:related_work}

Dataset Condensation aims to create a small, efficient dataset to reduce the cost of training large models. The field has evolved from selecting subsets of real data to synthesizing entirely new data points, with most efforts historically focused on uni-modal data.

\noindent \textbf{Uni-modal Data Condensation}. Uni-modal condensation methods fall into two main categories. \textit{Data Selection} (or coreset selection) identifies a representative subset of the original data based on various criteria like clustering~\citep{alexey2016discriminative, bautista2016cliquecnn}, greedy utility maximization~\citep{wei2015submodularity, soper2021greed}, gradient information~\citep{paul2021deep,mirzasoleiman2020coresets,killamsetty2021grad,LESS}, or other model-based metrics~\citep{toneva2018empirical, wang2025datawhisperer,wang2025winning,wang2025circuitseer}. However, these methods are fundamentally limited by being confined to the original data sources and cannot yield unseen data. In contrast, \textit{Dataset Distillation}~\citep{DD} synthesizes a small set of new, optimized data points. The goal is to match the learning dynamics of the full dataset. Dominant strategies include \textit{Gradient Matching}, which aligns training gradients~\citep{DC,DCC,DSA,wang2024samples,wang2025drupi} or entire parameter trajectories~\citep{DATM, MTT}, and \textit{Distribution Matching}, which aligns feature statistics in a pretrained embedding space~\citep{DM, IDM, NCFM}. Our work builds upon this data synthesis paradigm, adapting it for the unique challenges of the multi-modal setting.

\noindent \textbf{Multi-modal Data Condensation}. Extending condensation to multi-modal data is a nascent but critical research direction. The primary challenge is not only preserving the intra-modal statistics but, more importantly, the cross-modal semantic relationships. Previous works focus on multi-modal data selection A few recent methods have begun to tackle this problem. Although recent advancements in data selection, such as LLM-based filtering~\cite{chen2023alpagasus, liu2023makes, xu2023rethinking}, gradient-based influence estimation~\cite{attendu2023nlu}, and self-instruction generation~\cite{kung2023active}, have shown promise in optimizing instruction tuning, their effectiveness can be inconsistent, with some studies indicating they often fail to consistently outperform random sampling. AVDD~\citep{kushwaha2024audio} performs condensation for audio-visual data but does so by matching distributions in separate, modality-specific feature spaces, which risks misalignment and relies on now-outdated backbone architectures. To address cross-modal relationships, LoRS~\citep{xu2024low} proposes matching a pre-computed ground-truth similarity matrix between modalities, using low-rank factorization for efficiency. However, boiling down the complex, high-dimensional relationship between modalities to a single scalar similarity may be an oversimplification that fails to capture the full distributional structure. More recently, RepBlend~\citep{zhang2025beyond} introduced representation blending to encourage diversity and prevent modality collapse, but this remains a heuristic approach that can be difficult to balance and may not guarantee the preservation of joint-modal semantics.


\section{Methodology}

\subsection{Problem Formulation}
Dataset Condensation aims to obtain a small, information-rich dataset $\mathcal{S}$ that acts as an efficient substitute for a large real dataset $\mathcal{R}$. Normally, $\mathcal{S}$ consists of a rather small DPC. Formally, given a real dataset $\mathcal{R} = \{(x_i, y_i)\}_{i=1}^N$, the goal is to generate a much smaller (synthetic) dataset $\mathcal{S} = \{(\tilde{x}_j, y_j)\}_{j=1}^M$ where $M \ll N$. While many works frame this as a complex bi-level optimization problem involving unrolled model training~\citep{MTT, DSA}, a more efficient and increasingly popular paradigm is to perform \textit{distribution matching} in a fixed feature space. This approach avoids the expensive inner-loop model training. Let $\psi$ be a feature extractor, which remains frozen during condensation, mapping an arbitrary input $x$ into $e_x=\psi(x)$. The optimization problem simplifies to a single-level objective:
\begin{equation}
\label{eq:single_level}
\min_{\mathcal{S}} \mathbf{D}_{x\sim \mathcal{R}, \tilde{x}\sim \mathcal{S}}\big(e_x, e_{\tilde{x}}\big),
\end{equation}
where $x$ and $\tilde{x}$ denote the sets of feature embeddings for the real and synthetic data, and $\mathbf{D}(\cdot, \cdot)$ is a distributional distance metric. The core challenge, which we address, is to design an effective metric $\mathbf{D}$ for the multi-modal setting.

\subsection{Distribution Matching}
A common choice for the distance metric $\mathbf{D}$ is the Maximum Mean Discrepancy (MMD)~\citep{DM,IDM}. MMD measures the distance between two distributions, real data distribution $P_{\mathcal{R}}$ and synthetic data distribution $P_{\mathcal{S}}$, by mapping them into a empirical kernel space and computing the distance between their mean embeddings. While widely used, the effectiveness of MMD is highly dependent on the choice of the kernel, which is often heuristic (\emph{e.g.}, a Gaussian kernel). A poorly chosen kernel may fail to capture all statistical differences between the distributions, leading to suboptimal alignment.

To address the limitations of kernel-based methods, we adopt a more principled metric founded on the Characteristic Function (CF)~\citep{NCFM,NCFGAN}. The CF of a random vector $z \in \mathcal{Z}$ is the Fourier transform of its probability density function (PDF), which uniquely defines its probability distribution. It is defined as:
\begin{equation}
    \Phi(z;t) = \mathbb{E}_{z \sim P_\mathcal{Z}} \left[ e^{j t^\top z} \right],
\end{equation}
where $j = \sqrt{-1}$ is the imaginary unit and $t \in \mathbb{R}^{\dim(\mathcal{Z})}$ is a frequency vector. By Lévy's Uniqueness Theorem, two distributions are identical if and only if their characteristic functions are identical. This allows us to define the \textit{Characteristic Function Discrepancy} (CFD) as the squared $L_2$-distance between the CFs of the real and synthetic feature distributions, $P_{\mathcal{R}}$ and $P_{\mathcal{S}}$:
\begin{equation}\label{eq:chfloss}
\begin{aligned}
 &\text{CFD}(x,\tilde{x}) = \big(\Phi(x;t) -\Phi(\tilde{x};t)\big)\big(\bar\Phi(x;t) -\bar\Phi(\tilde{x};t)\big) \\
   & =  |\Phi(x;t)|^2 + |\Phi({\tilde{x}};t)|^2 \\
    & - |\Phi(x;t)| |\Phi({\tilde{x}};t)| \left(2 \cos(a_x(t) - a_{\tilde{x}}(t)\right) \\
    & = |\Phi(x;t) - \Phi({\tilde{x}};t)|^2 \\
    & + 2 |\Phi(x;t)| |\Phi({\tilde{x}};t)| \left( 1 - \cos(a_x(t) - a_{\tilde{x}}(t)) \right),
\end{aligned}
\end{equation}
where the expectation is taken over a distribution of random frequency vectors $t$. Empirically, we approximate the CFs and the expectation with samples. Unlike MMD, CFD does not depend on a user-defined kernel and provides a more robust framework for matching distributions by comparing all of their statistical moments implicitly in the Fourier domain~\citep{NCFM}.

\subsection{ImageBindDC Framework}
We now introduce our multi-modal data condensation framework, ImageBindDC, as illustrated in Figure~\ref{fig:ImageBindDC}. Without loss of generality, we take \textit{audio and vision} multi-modal data as an example to illusate ImageBindD framework. The goal of ImageBindDC is to learn a small multi-modal synthetic dataset, $\mathcal{S} = \mathcal{S}_a \cup \mathcal{S}_v$, that efficiently represents a much larger real dataset, $\mathcal{R} = \mathcal{R}_a \cup \mathcal{R}_v$. As shown in Figure~\ref{fig:ImageBindDC}(a), our method leverages a pretrained ImageBind encoder, which provides a unified embedding space for different modalities. Both the real data instance ($x_a$, $x_v$), and the synthetic data instance ($\tilde{x}_a$, $\tilde{x}_v$), are passed through the ImageBindDC pipeline to obtain their respective embeddings, ($e_a, e_v$) and ($\tilde{e}_a, \tilde{e}_v$), within this common space.

The core of ImageBindDC is to minimize the discrepancy between the distribution of real embeddings and synthetic embeddings for all modalities, as shown in (Figure~\ref{fig:ImageBindDC}b). By aligning the distributions in this shared space, ImageBindDC ensures that the condensed synthetic data captures the essential statistical characteristics of the original, large-scale multi-modal dataset.

\noindent \textbf{Multi-modal Modeling}. In our approach, we leverage ImageBind~\citep{Imagebind}, which is designed to map multi-modal data into a shared feature space. The core idea behind ImageBind is to create a unified embedding space where different types of data can be represented as points in the same space, facilitating effective multi-modal learning. ImageBind operates by learning a set of embeddings for each modality, such as images and audio, and binding them into a shared feature space. It ensures that each modality is represented in a way that reflects its unique characteristics while also aligning the modalities within a shared space.

Specifically, let $ e_{a} $ and $ e_{v} $ represent the embeddings for audio and image data, respectively. These embeddings are learned such that they can be projected into a feature space $ \mathcal{F} $. The objective of the ImageBind is to minimize the distance between corresponding audio and image embeddings within the shared space. This is typically achieved by optimizing a loss function that ensures that similar data points from different modalities are aligned in the feature space. The embedding process for each modality is formulated as:
\begin{equation}
e_{a} = \mathcal{E}_{a}(x_{a}), \quad e_{v} = \mathcal{E}_{v}(x_{v}),
\end{equation}
where $\mathcal{E}_{a}$ and $ \mathcal{E}_{v} $ are the encoding functions for audio and image data, respectively, and $ x_{a} $ and $ x_{v} $ are the input audio and image data points.

\noindent \textbf{Uni-modal Alignment}. In the context of uni-modal distribution matching, we aim to minimize the discrepancy between the embeddings of real and synthetic data for each modality separately. This is achieved by employing the CFD in Eq.~(\ref{eq:chfloss}), which is designed to align the distribution of embeddings from real data with that of synthetic data in the shared feature space.

Specifically, for audio data, let $ e_{a, \mathrm{ real}} $ and $ e_{a} $ denote the embeddings of real and synthetic audio data, respectively. Similarly, for image data, let $ e_{v, \mathrm{real}} $ and $ e_{v, \mathrm{syn}} $ represent the embeddings of real and synthetic images, respectively. The objective of Uni-modal Distribution Matching is to minimize the embedding discrepancy for each modality, as shown in Figure~\ref{fig:ImageBindDC}(c)i. Specifically, we define the CF loss for the audio modality as follows:
\begin{equation}
\mathcal{L}_{\mathrm{audio}} = \mathrm{CFD}(e_a,\tilde{e}_a).
\end{equation}
The CF loss computes the difference between the characteristic functions of the real and synthetic audio embeddings, measured in terms of the CFD. Similarly, for image data:
\begin{equation}
\mathcal{L}_{\mathrm{image}} = \mathrm{CFD}(e_v,\tilde{e}_v).
\end{equation}
The total loss for the Uni-modal Distribution Matching across both modalities is the sum of the individual losses:
\begin{equation}
\mathcal{L}_{\mathrm{uni}} = \mathcal{L}_{\mathrm{audio}} + \mathcal{L}_{\mathrm{image}}.
\end{equation}
By minimizing this loss, we ensure that the embeddings of real and synthetic data for both audio and image modalities are aligned in the  feature space, facilitating better distribution matching guaranteed by CFD's principled properties.

\noindent \textbf{Cross-modal Alignment}. In the Cross-modal Alignment (CMA), we aim to measure the alignment between the real audio and image embeddings, as well as between the synthetic audio and image embeddings. Specifically, we perform element-wise multiplication between the embeddings of real audio and image data, and similarly for synthetic audio and image embeddings. The resulting values are then used to compute the cosine similarity between the real and synthetic data embeddings.

Specifically, let $e_{a}$ and $e_{v}$ represent the real audio and image embeddings, respectively, while $\tilde{e}_{a}$ and $
\tilde{e}_{v} $ represent the synthetic audio and image embeddings, respectively. The element-wise multiplication of the embeddings for real audio and image data is defined as:
\begin{equation}
e_{a} \odot e_{v} = \left[ e_{a, 1} \cdot e_{v, 1}, \dots, e_{a, N} \cdot e_{v, N} \right],
\end{equation}
where $ \odot $ denotes element-wise multiplication and $ N $ is the dimension of the embedding vectors. Similarly, for synthetic audio and image data, the element-wise multiplication is defined as:
\begin{equation}
\tilde{e}_{a} \odot \tilde{e}_{v} = \left[ \tilde{e}_{a, 1} \cdot \tilde{e}_{v\, 1}, \dots, \tilde{e}_{a, N} \cdot \tilde{e}_{v\, N} \right].
\end{equation}
Next, we compute the cosine similarity between the element-wise multiplied real and synthetic embeddings. The cosine similarity for the real embeddings is given by:
\begin{equation}
\rho_\mathrm{cross} = \frac{\langle e_{a} \odot e_{v}, \tilde{e}_{a} \odot \tilde{e}_{v} \rangle}{\|e_{a} \odot e_{v}\|_2 \|\tilde{e}_{a} \odot \tilde{e}_{v}\|_2},
\end{equation}
where $ \langle \cdot, \cdot \rangle $ represents the dot product and $ \|\cdot\|_2 $ denotes the L2 norm. The cosine similarity computes the alignment between the element-wise multiplied real and synthetic embeddings, with higher values indicating better alignment. The final Cross-modal Alignment (CMA) loss is then calculated by minimizing the cosine similarity for both real and synthetic data:
\begin{equation}
\mathcal{L}_{\mathrm{cross}} = 1 - \rho_\mathrm{cross}.
\end{equation}
By minimizing this loss, the model learns to align the real and synthetic audio-image embeddings, facilitating better cross-modal matching.

\noindent \textbf{Joint-Modal Alignment.} In the Joint-Modal Alignment (JMA) part, we focus on minimizing the gap between the average embeddings of real and synthetic data across different modalities. The key idea is to compute the mean embedding for each modality (audio and image) and then perform a matrix multiplication between the real and synthetic embeddings to calculate a cross-modal similarity score.

Specifically, let $ e_{a} $, $ \tilde{e}_{a} $, $ e_{v} $, and $ \tilde{e}_{v, \mathrm{syn}} $ represent the embeddings for real audio, synthetic audio, real image, and synthetic image data, respectively. First, we compute the mean of the embeddings along each dimension.
The mean embeddings for each modality (audio and image) are denoted by $ {E}_{a} $, $ {\tilde{E}}_{a} $, $ {E}_{v} $, and $ {\tilde{E}}_{v} $. Next, we reshape the average embeddings into 2D matrices for matrix multiplication, where each average embedding is transformed into a row vector. Then, we compute the cross-modal similarity score by performing a matrix multiplication between the element-wise multiplied average embeddings:
\begin{equation}
\rho_\mathrm{joint} = {E}_{a} \odot \tilde{E}_{v}^\top \times {E}_{v} \odot \tilde{E}_{a}^\top
\end{equation}
where $ \times $ denotes the matrix multiplication operation. This step computes the similarity between the real and synthetic embeddings, considering the alignment between both modalities. Finally, the JMA loss is calculated as the mean of the joint-modal gap, which is given by:
\begin{equation}
\mathcal{L}_{\mathrm{joint}} = 1 - \rho_\mathrm{joint},
\end{equation}
This loss encourages the model to minimize the gap between real and synthetic multi-modal embeddings, thereby enhancing the alignment between the two modalities.

\noindent \textbf{Put All Components Together}. The final loss function is obtained by combining the above mentioned three distinct losses: $ \mathcal{L}_{\mathrm{uni}} $, $ \mathcal{L}_{\mathrm{cross}} $, and $ \mathcal{L}_{\mathrm{joint}} $, each corresponding to different levels of alignment between the real and synthetic data distributions. These losses are scaled by respective hyperparameters $ \lambda_{\mathrm{uni}} $, $ \lambda_{\mathrm{cross}} $, and $ \lambda_{\mathrm{joint}} $, which control their contribution to the overall objective. The final loss is defined as:
\begin{equation}
\mathcal{L} = \lambda_{\mathrm{uni}} \mathcal{L}_{\mathrm{uni}} + \lambda_{\mathrm{cross}} \mathcal{L}_{\mathrm{cross}} + \lambda_{\mathrm{joint}} \mathcal{L}_{\mathrm{joint}}.
\end{equation}
The detailed pseudocode of the overall framework including input and output is provided in Appendix~\ref{alg:imagebinddc}.

\section{Experiments}

\begin{table}[tb!]
\caption{Classification accuracy (\%) for data condensation on VGGS-10K and AVE under different DPC settings. All methods use a randomly initialized ConvNet to guide distillation, with accuracy measured by training that same ConvNet from scratch on the resulting condensed data.}
\label{tab:main_conv}
\resizebox{0.99\textwidth}{!}{
\begin{tabular}{@{}ccc|cccc|cccccc|c@{}}
\toprule
 &  &  & \multicolumn{4}{|c|}{Selection-based} & \multicolumn{6}{c|}{ Distillation-based} & \multirow{2}{*}{Whole data} \\ 
Dataset & DPC & Ratio \% & Random & Herding & Forgetting & GraNd & DC & DSA & MTT & DM & AVDD & \textbf{ImageBindDC}  \\ \midrule
\multirow{3}{*}{VGGS-10K} & 1 & 0.11 & 15.44{\scriptsize ±1.87} & 20.77{\scriptsize ±2.11} & 23.41{\scriptsize ±1.31} & 15.42{\scriptsize ±0.42} & 18.28{\scriptsize ±1.36} & 19.32{\scriptsize ±1.35} & 34.13{\scriptsize ±3.6} & 36.54{\scriptsize ±2.52} & 40.41{\scriptsize ±1.81} & \textbf{42.66{\scriptsize ±1.48}} & \multirow{3}{*}{68.24{\scriptsize ±0.75}} \\
 & 10 & 1.13 & 32.01{\scriptsize ±1.64} & 39.89{\scriptsize ±1.64} & 40.78{\scriptsize ±2.04} & 34.95{\scriptsize ±1.52} & 32.10{\scriptsize ±0.84} & 36.61{\scriptsize ±1.04} & 36.79{\scriptsize ±1.97} & 43.85{\scriptsize ±1.75} & 48.08{\scriptsize ±0.92} & \textbf{55.23{\scriptsize ±0.13}} &  \\
 & 20 & 2.26 & 45.1{\scriptsize ±2.31} & 50.2{\scriptsize ±0.74} & 52.16{\scriptsize ±0.49} & 49.22{\scriptsize ±1.22} & - & - & 51.87{\scriptsize ±1.26} & 49.01{\scriptsize ±2.44} & 48.86{\scriptsize ±1.53} & \textbf{55.30{\scriptsize ±0.18}} &   \\ \midrule
\multirow{3}{*}{AVE} & 1 & 0.1 & 10.07{\scriptsize ±1.16} & 11.84{\scriptsize ±0.4} & 10.07{\scriptsize ±0.45} & 8.69{\scriptsize ±0.42} & 10.45{\scriptsize ±0.39} & 10.76{\scriptsize ±0.62} & 12.13{\scriptsize ±0.41} & 16.70{\scriptsize ±1.46} & 16.90{\scriptsize ±0.14} & \textbf{18.08{\scriptsize ±0.52}} & \multirow{3}{*}{52.20{\scriptsize ±0.38}} \\
 & 10 & 1 & 13.64{\scriptsize ±0.22} & 21.94{\scriptsize ±0.52} & 20.31{\scriptsize ±0.38} & 19.54{\scriptsize ±0.35} & 22.04{\scriptsize ±1.04} & 20.92{\scriptsize ±1.00} & 23.15{\scriptsize ±0.95} & 26.14{\scriptsize ±1.80} & 32.90{\scriptsize ±0.14} & \textbf{34.42{\scriptsize ±0.32}} &  \\
 & 20 & 2 & 26.32{\scriptsize ±1.01} & 33.04{\scriptsize ±0.38} & 31.17{\scriptsize ±0.49} & 29.27{\scriptsize ±0.51} & - & - & - & 32.57{\scriptsize ±0.97} & 36.67{\scriptsize ±0.49} & \textbf{38.41{\scriptsize ±0.07}} &  \\ \bottomrule
\end{tabular}
}
\end{table}

\begin{table}[tb!]
\caption{Classification accuracy (\%) for data condensation methods on VGGS-10K and AVE under different DPC settings. The condensation for all methods is guided by a pretrained ImageBind model, with accuracy measured by training that same model from scratch on the condensed data. Note that OOM indicates that the method ran out of memory on 2 H100 GPUs.}
\label{tab:main_imagebind}
\resizebox{0.99\textwidth}{!}{
\begin{tabular}{@{}ccc|cccc|cccccc|c@{}}
\toprule
& & & \multicolumn{4}{|c|}{Selection-based} & \multicolumn{6}{c|}{ Distillation-based} & \multirow{2}{*}{Whole data} \\
Dataset & DPC & Ratio (\%) & Random & Herding & Forgetting & GraNd & DC & DSA & MTT & DM & AVDD & \textbf{ImageBindDC}  \\ \midrule
\multirow{3}{*}{VGGS-10K} & 1 & 0.11 & 25.24{\scriptsize ±3.25} & 27.14{\scriptsize ±1.58} & 29.34{\scriptsize ±1.49} & 25.33{\scriptsize ±2.03} & OOM & 28.87{\scriptsize ±1.50} & OOM & 31.55{\scriptsize ±1.34} & 36.10{\scriptsize ±1.67} & \textbf{37.25{\scriptsize ±1.62}} & \multirow{3}{*}{60.57{\scriptsize ±0.05}} \\
 & 10 & 1.13 & 40.86{\scriptsize ±1.32} & 43.15{\scriptsize ±1.28} & 44.67{\scriptsize ±1.07} & 41.06{\scriptsize ±1.13} & OOM & 44.73{\scriptsize ±1.36} & OOM & 45.46{\scriptsize ±1.27} & 46.61{\scriptsize ±0.59} & \textbf{48.89{\scriptsize ±0.78}} & \\
& 20 & 2.26 & 50.18{\scriptsize ±1.66} & 51.92{\scriptsize ±0.98} & 52.98{\scriptsize ±0.79} & 50.77{\scriptsize ±1.09} & OOM & 52.72{\scriptsize ±1.37} & OOM & 53.17{\scriptsize ±0.89} & 53.87{\scriptsize ±1.14} & \textbf{56.11{\scriptsize ±0.98}} &  \\ \midrule
\multirow{3}{*}{AVE} & 1 & 0.1 & 45.11{\scriptsize ±3.17} & 45.77{\scriptsize ±1.75} & 43.23{\scriptsize ±1.88} & 40.62{\scriptsize ±2.08} & OOM & 44.46{\scriptsize ±2.16} & OOM & 65.35{\scriptsize ±2.06} & 67.32{\scriptsize ±1.55} & \textbf{70.10{\scriptsize ±1.24}} & \multirow{3}{*}{76.93{\scriptsize ±0.14}} \\
& 10 & 1 & 64.56{\scriptsize ±1.88} & 66.76{\scriptsize ±1.24} & 65.25{\scriptsize ±1.45} & 62.06{\scriptsize ±1.19} & OOM & 58.49{\scriptsize ±5.05} & OOM & 69.49{\scriptsize ±0.69} & 71.33{\scriptsize ±0.35} & \textbf{73.67{\scriptsize ±0.31}} & \\
& 20 & 2 & 67.01{\scriptsize ±1.96} & 69.90{\scriptsize ±0.47} & 68.75{\scriptsize ±0.64} & 65.66{\scriptsize ±0.76} & OOM & 71.49{\scriptsize ±0.51} & OOM & 70.50{\scriptsize ±0.29} & 72.81{\scriptsize ±0.24} & \textbf{75.34{\scriptsize ±0.27}} & \\ \bottomrule
\end{tabular}
}
\end{table}

\begin{table}[tb!]
\caption{Accuracy (\%) on the NYU-v2 dataset for a depth-text classification task. The text modality is derived from the scene name (\emph{e.g.}, ``bathroom''). For this experiment, a randomly initialized ImageBind model guides the condensation, and performance is evaluated by training the same model architecture from scratch on the condensed data.}
\label{tab:depth_text}
\centering
\resizebox{0.6\textwidth}{!}{
\begin{tabular}{@{}c|cccc@{}}
\toprule
DPC & 1 & 2 & 5 & 10 \\ 
Ratio (\%) & 0.034 & 6.79 & 16.98 & 33.96 \\ \midrule
Random & 60.60{\scriptsize ±2.43} & 69.22{\scriptsize ±1.87} & 73.85{\scriptsize ±1.42} & 88.38{\scriptsize ±3.05} \\
DM & 67.97{\scriptsize ±11.69} & 75.25{\scriptsize ±12.92} & 89.08{\scriptsize ±4.96} & 96.89{\scriptsize ±1.26} \\
AVDD & 72.22{\scriptsize ±10.03} & 81.30{\scriptsize ±2.84} & 95.92{\scriptsize ±1.63} & 98.62{\scriptsize ±0.45} \\
\textbf{ImageBindDC} & \textbf{80.43{\scriptsize ±0.44}} & \textbf{88.33{\scriptsize ±4.99}} & \textbf{97.30{\scriptsize ±1.17} }& \textbf{ 98.73{\scriptsize ±1.04} }\\
Whole Data& \multicolumn{4}{c}{98.62{\scriptsize ±0.25}} \\
\bottomrule
\end{tabular}
}
\end{table}

\begin{table}[tb!]
\centering
\caption{Audio-text retrieval performance (Recall@K) on the Clotho dataset. All methods distill features from the pre-trained ImageBind space at a setting of 20 DPC. Performance is evaluated by training an ImageBind model, which comprises a frozen backbone and a trainable linear head.}
\label{tab:clotho_retrieval_results}
\resizebox{0.8\textwidth}{!}{
\begin{tabular}{@{}c|ccc|ccc@{}}
\toprule
\multirow{2}{*}{Metric} & \multicolumn{3}{c|}{A2T} & \multicolumn{3}{c}{T2A} \\ 
 & R@1 & R@5 & R@10 & R@1 & R@5 & R@10 \\ \midrule
DM & 0.0268 & 0.1014 & 0.1761 & 0.0306 & 0.1148 & 0. 1684 \\
AVDD & 0.0316 & 0.1024 & 0.1866 & 0.0402 & 0.133 & 0.2048 \\
\textbf{ImageBindDC} &\textbf{ 0.0362} & \textbf{0.1297} &\textbf{ 0.1995} & \textbf{0.0464} & \textbf{0.1567} & \textbf{0.2281} \\
Whole Data & 0.0526 & 0.1627 & 0.2364 & 0.0565 & 0.1674 & 0.2488 \\
\bottomrule
\end{tabular}
}
\end{table}

\subsection{Experimental Settings}

\noindent \textbf{Audio-visual Tasks}. For audio-visual tasks, following~\citep{zhao2023dataset, kushwaha2024audio, cazenavette2022dataset}, we evaluated all distillation methods using two widely used audio-visual datasets: VGGS-10K and AVE. VGGSound~\citep{chen2020VGGSound} is a large-scale audio-visual dataset containing approximately 200k YouTube videos across 309 classes. VGGS-10K\cite{VGG} is a subset of VGGSound. AVE~\citep{tian2018audio} consists of 4,143 video clips spanning 28 event categories. 

\noindent \textbf{Text-image Tasks}. We utilized NYU-v2~\cite{NYU} dataset. The text modality contains the name of the scene from which the corresponding image was captured. This dataset includes RGB and depth images. 

\noindent \textbf{Audio-text Tasks}. Experiments were conducted on Clotho~\citep{Clotho}, a large-scale audio captioning dataset designed for training and evaluating models that generate captions for audio clips.

\noindent \textbf{Baselines}. For selection-based methods, we utilized Random, Herding~\citep{welling2009herding}, Forgetting~\citep{toneva2018empirical}, GraNd~\citep{paul2021deep}. For distillation-based methods, we included DC~\citep{DC}, DSA~\citep{DSA}, MTT~\citep{MTT}, DM~\citep{DM} and AVDD~\citep{kushwaha2024audio}. More details of experimental settings are provided in Appendix~\ref{app:details}.

\begin{table}[tb!]
\centering
\caption{Cross-architecture accuracy (\%) on the AVE dataset. A pretrained ImageBind model is used to guide the distillation of synthetic datasets, which are then evaluated by training different architectures from scratch.}
\label{tab:cross_arch_imagebind}
\resizebox{0.7\textwidth}{!}{
\begin{tabular}{@{}c|cc|cc@{}}
\toprule
Condensation  & \multicolumn{2}{c|}{1 DPC} & \multicolumn{2}{c}{10 DPC} \\ 
Ratio (\%) & \multicolumn{2}{c|}{0.1} & \multicolumn{2}{c}{1} \\ \midrule
Model & ConvNet & ImageBind & ConvNet & ImageBind \\
Random & 10.07{\scriptsize ±1.16} & 45.11{\scriptsize ±3.17} & 13.64{\scriptsize ±0.22} & 64.56{\scriptsize ±1.88} \\
MTT & OOM & OOM & OOM & OOM \\
DM & 7.10{\scriptsize ±1.36} & 60.51{\scriptsize ±1.12} & 11.61{\scriptsize ±0.88} & 69.26{\scriptsize ±1.67} \\
AVDD & 4.85{\scriptsize ±1.11} & 67.32{\scriptsize ±1.55} & 12.44{\scriptsize ±1.29} & 71.33{\scriptsize ±0.35} \\
\textbf{ImageBindDC} & \textbf{12.69{\scriptsize ±1.23}} & \textbf{70.10{\scriptsize ±1.24}} & \textbf{16.74{\scriptsize ±0.18}} & \textbf{73.67{\scriptsize ±0.31}} \\ \midrule
Whole & 52.20{\scriptsize ±0.38} & 76.93{\scriptsize ±0.14} & 52.20{\scriptsize ±0.38} & 76.93{\scriptsize ±0.14} \\ 
\bottomrule
\end{tabular}
}
\end{table}

\noindent \textbf{Models}. We employed two distinct architectures for encoding the inputs: ImageBind and a ConvNet architecture. ImageBind~\cite{Imagebind} is used to map both audio and image data into a shared feature space.

\noindent \textbf{Implementation Details}. Experiments were all conducted on 4 NVIDIA A100 GPUs. For AVE and VGGS-10K, we used a learning rate of 0.2 and an SGD optimizer with a momentum of 0.5. NYU-v2, we used a learning rate of 0.001 for both the backbone and the classifier, with the same optimizer. To ensure fairness, all experiments were conducted for 3 times. More details is included in Appendix~\ref{app:details}.

\begin{table}[tb!]
\centering
\caption{Computational efficiency of ImageBindDC and different methods on VGGS-10K. Note that OOM indicates that the method ran out of memory on a single A100 GPU.}
\label{tab:efficiency}
\resizebox{0.8\textwidth}{!}{
\centering
\begin{tabular}{@{}c|ccc|ccc@{}}
\toprule
 Metric & \multicolumn{3}{c|}{Condensation Time (s)} & \multicolumn{3}{c}{GPU Memory (GB)} \\ 
DPC & 1 & 10 & 20 & 1 & 10 & 20 \\ \midrule
DSA & 208.48 & 508.36 & 712.01 & 9.13 & 9.36 & 14.15 \\
DM & 140.3 & 611.15 & 707.21 & 8.96 & 9.34 & 14.24 \\
MTT & OOM & OOM & OOM & OOM & OOM & OOM \\
DC & OOM & OOM & OOM & OOM & OOM & OOM \\
AVDD & 158.11 & 507.08 & 700.1 & 8.96 & 9.34 & 14.24 \\
\textbf{ImageBindDC} & \textbf{57.46} & \textbf{89.4} & \textbf{123.74} & \textbf{5.6} & \textbf{8.14} & \textbf{13.39} \\  \bottomrule
\end{tabular}
}
\end{table}

\subsection{Main Results}
\noindent \textbf{Comparison with Baselines}. 
Our ImageBindDC approach consistently outperforms other methods across all datasets and settings. As shown in Table~\ref{tab:main_conv} and~\ref{tab:main_imagebind}, ImageBindDC achieves the highest classification accuracy on both VGGS-10K and AVE datasets, under different DPC settings, when using both ConvNet and ImageBind models for distillation. For instance, on VGGS-10K with 10 DPC and using the ImageBind model, ImageBindDC achieves an accuracy of 55.23\%, a significant improvement over the next best method, AVDD (48.08\%). Similarly, on the NYU-v2 dataset shown in Table~\ref{tab:depth_text}, ImageBindDC demonstrates superior performance, achieving 98.73\% accuracy with 10 DPC, while the baseline DM method reaches 96.89\%. Furthermore, as shown in Table~\ref{tab:clotho_retrieval_results} ImageBindDC also have outstanding performance on audio-text retrieval.

\noindent \textbf{Cross-Architecture Performance}. We tested the performance of the data distilled by our methods on unseen models. As shown in Table~\ref{tab:cross_arch_imagebind}, a pretrained ImageBind model is used to guide the distillation of synthetic datasets, which are then evaluated by training different architectures from scratch, \emph{e.g.}, Convnets and ImageBind. ImageBindDC consistently outperforms other methods. For example, with 10 DPC, ImageBindDC achieves 14.42\% accuracy on ConvNet and 73.67\% on ImageBind, largely surpassing previous SOTA (12.44\% on ConvNet and 71.33\% on ImageBind).

\begin{figure}[htbp]
    \centering
    \includegraphics[width=0.7\linewidth]{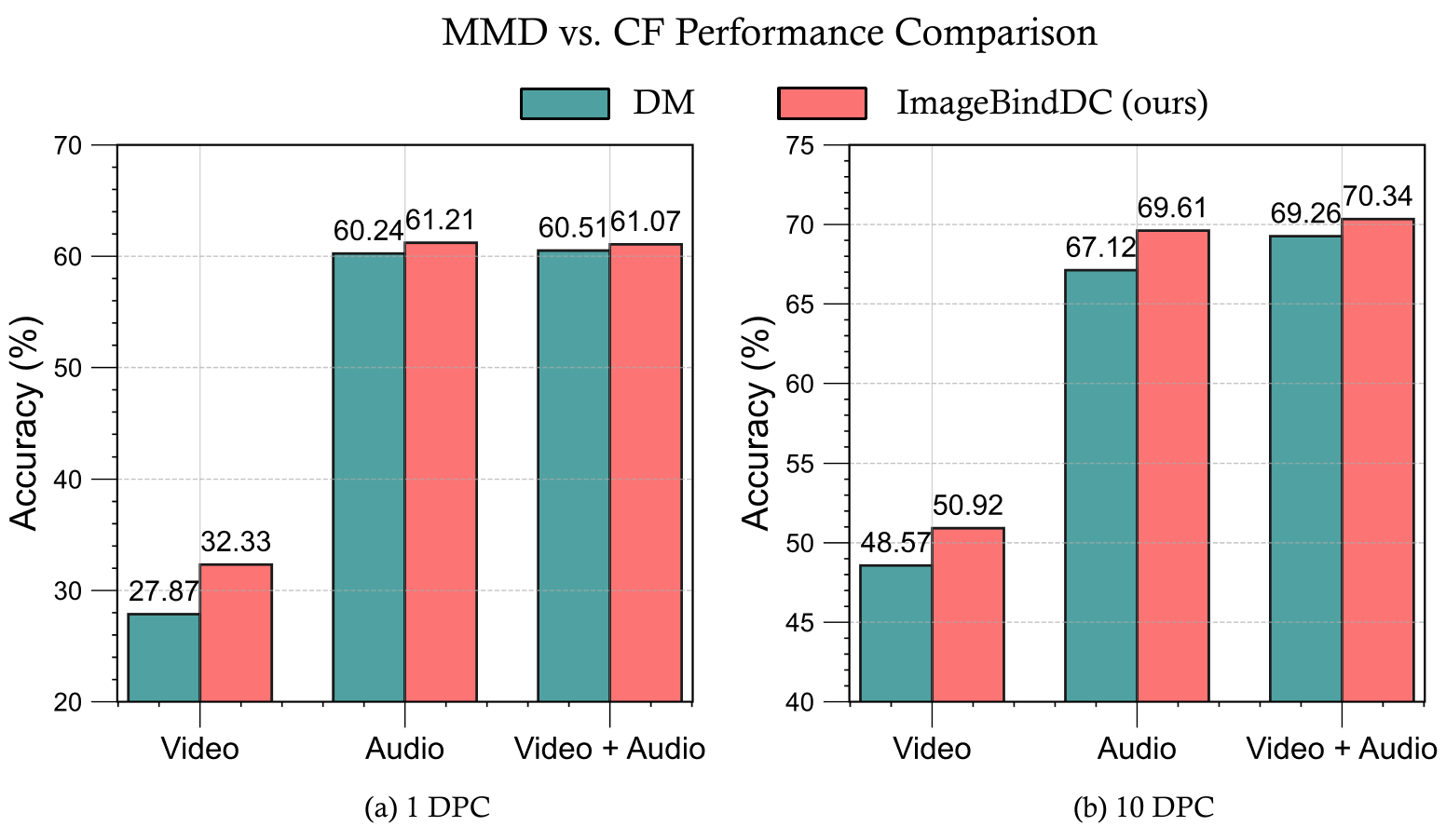}
    \caption{DM (MMD) vs. ImageBindDC (CF) Performance Comparison. The figure illustrates the accuracy of the MMD and CF methods under (a) 1 DPC and (b) 10 DPC. In all matching configurations, including Video-only, Audio-only, and combined Video + Audio, ImageBindDC demonstrates superior performance over DM.}
    \label{fig:MMD_CF}
\end{figure}

\subsection{Ablation Study}
\noindent \textbf{Effectiveness of Characteristic Function Discrepancy (CFD)}. To validate our choice of a Characteristic Function Discrepancy (CFD), we compared its performance against the Maximum Mean Discrepancy (MMD) loss used by the DM baseline. Results in Figure~\ref{fig:MMD_CF} reveal a consistent advantage for our CF-based approach across all tested configurations. For uni-modal audio task with 1 DPC, ImageBindDC achieves 32.33\% accuracy, a significant lead over the 27.87\% from DM. This trend continues in the 10 DPC setting, where for the combined Video-Audio task, our method scores 70.34\% compared to DM's 69.26\%. While the improvements vary, the consistent superiority of our method in every scenario confirms that the CF loss provides a more robust and precise  alignment, making it a better-suited metric for the complexities of data condensation.

\begin{figure}[htbp]
    \centering
    \includegraphics[width=0.7\linewidth]{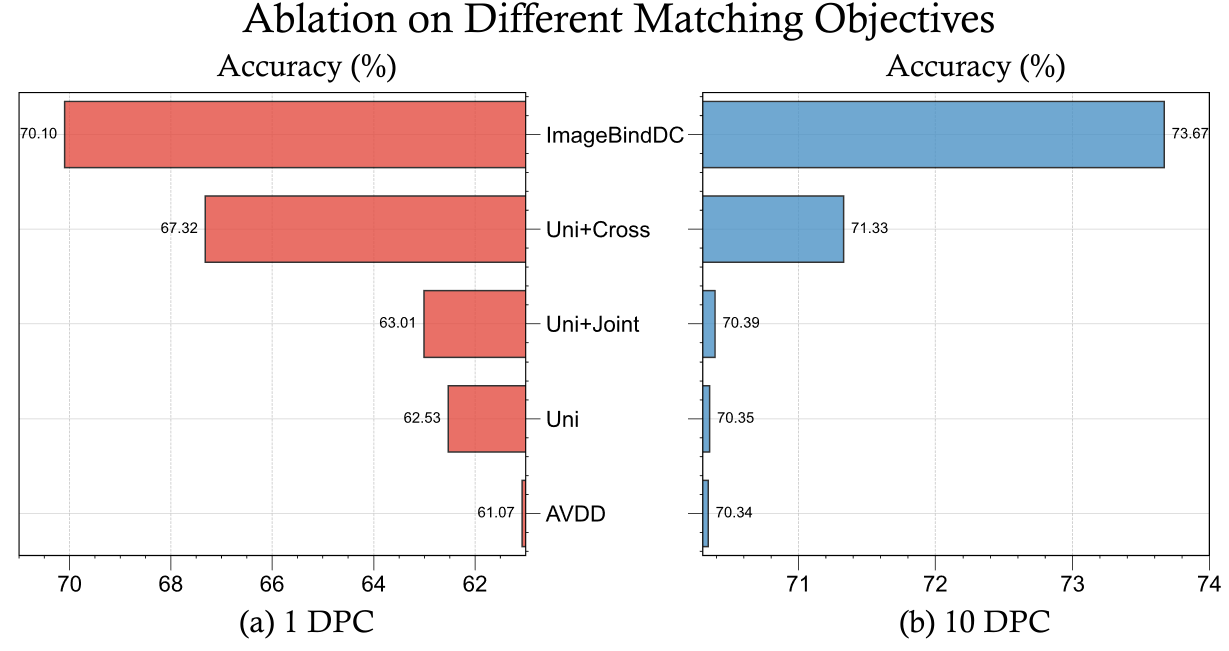}
    \caption{Ablation on Different Matching Objectives. This figure illustrates the contribution of uni-modal, joint-modal, and cross-modal matching objectives to overall accuracy. Results are presented for both (a) 1 DPC and (b) 10 DPC. The full configuration (ImageBindDC), combining all objectives, yields the best performance.}
    \label{fig:ablation}
\end{figure}

\noindent \textbf{Impact of Alignment Components}. To dissect the contribution of each alignment component in our proposed loss, we conducted further experiments, which are presented in Figure~\ref{fig:ablation}. The findings reveal that while preserving \textit{uni-modal} statistics provides a strong baseline, it is the combination of all three objectives that unlocks the full potential of the model. For instance, at 10 DPC, the \textit{uni-modal} loss alone achieves a respectable 70.34\% accuracy. Adding either the \textit{joint-modal} or \textit{cross-modal} objectives individually yields gains compared with the baseline. When all three components are combined, performance jumps significantly to 73.67\%, an absolute improvement of 3.33\%. This highlights a critical insight: \textit{\underline{alignment objectives are not merely additive but work in synergy}}. The \textit{uni-modal} loss preserves the integrity of each modality, while the \textit{cross-modal} and \textit{joint-modal} losses enforce the relational structure between them. Both are essential for creating a high-quality condensed dataset.

\noindent \textbf{Computational Efficiency Analysis}. We evaluated the computational efficiency of ImageBindDC against other baselines, with results presented in Table~\ref{tab:efficiency}. Our method is not only faster than competing condensation techniques but also dramatically reduces the resources required. For instance, at the 1 DPC setting, ImageBindDC is over 2.4$\times$ faster than DM (57.46s vs. 140.3s) and reduces GPU memory usage by a significant 37.5\% (5.6 GB vs. 8.96 GB). More critically, when compared to a single training epoch on the full dataset, the benefits are even more pronounced. Even at 20 DPC, our condensation process is over 3.4$\times$ faster than a single full-data epoch (123.74s vs. 419.9s) while slashing memory requirements by over 75\% (from 55.29 GB to 13.39 GB). This remarkable efficiency makes training large models feasible on resource-constrained systems.

\section{Discussion}
\begin{figure}[tb!]
    \centering
    \includegraphics[width=0.9\linewidth]{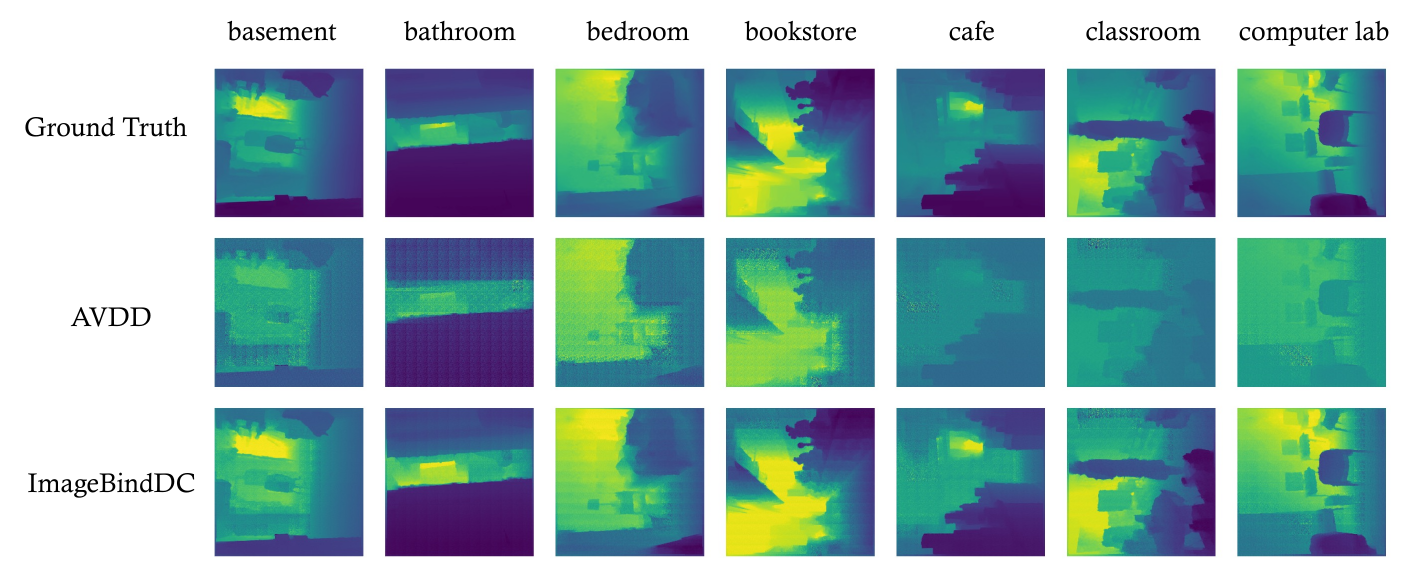}
    \caption{Qualitative comparison of distilled image samples on NYU-v2 dataset.  Images distilled by ImageBindDC demonstrate a superior ability to preserve the core visual coherence of the original data across all categories.}
    \label{fig:depth}
\end{figure}

\noindent \textbf{Visualization of distilled data}. Figure~\ref{fig:depth} provides a qualitative assessment of our method by visually comparing image samples from ImageBindDC, the AVDD baseline, and the original ground truth on the NYU-v2 dataset. The distilled images from ImageBindDC clearly demonstrate superior visual coherence and semantic integrity, with key features of scenes like 'bathroom' and 'bedroom' remaining easily recognizable. This stands in stark contrast to the baseline's samples, which often degrade into distorted or abstract amalgamations that fail to capture the defining characteristics of the original data. This qualitative advantage is a direct result of our core methodology; by performing condensation within a unified, joint-embedding space, ImageBindDC effectively preserves the intricate cross-modal relationships between visual data and their corresponding textual or depth information. This process prevents the semantic decoupling that plagues methods that condense modalities independently, yielding a synthetic dataset that is not only compact but also rich in high-fidelity features.

\begin{figure}[htbp]
    \centering
    \includegraphics[width=0.8\linewidth]{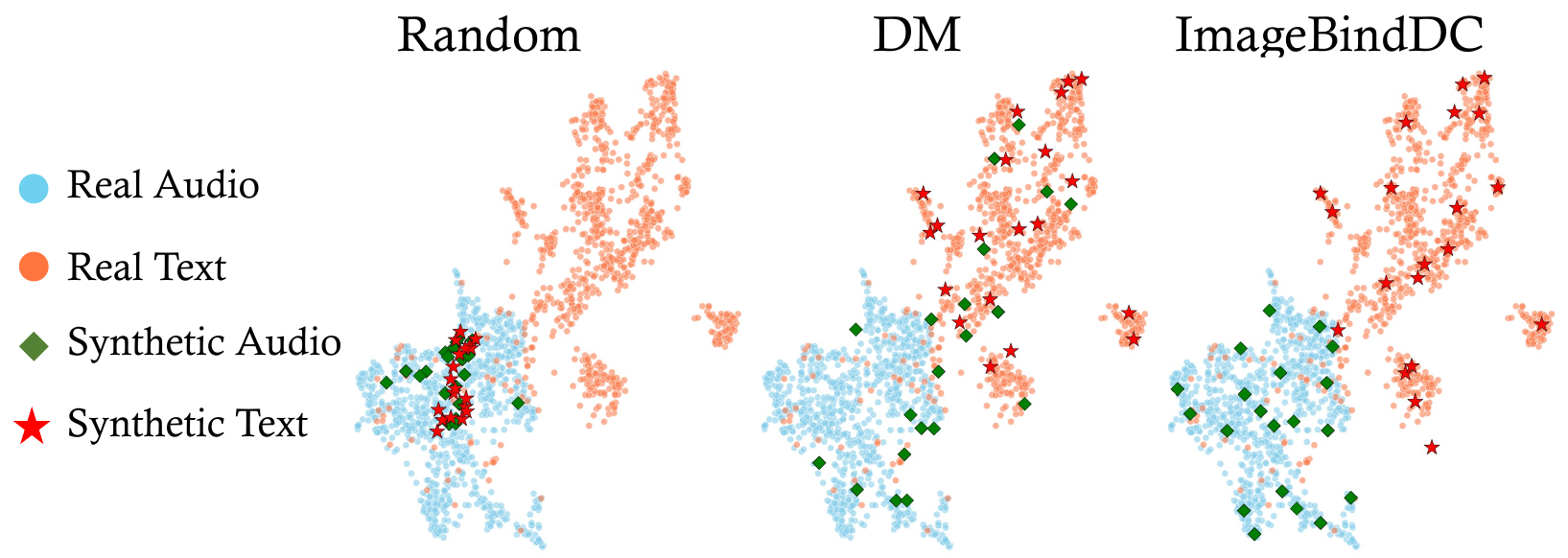}
    \caption{Umap of real and synthetic embeddings on Clotho dataset with audio and text modalities. It shows that ImageBindDC matches best to the real  data distribution.}
    \label{fig:embd}
\end{figure}

\noindent \textbf{Visualization of synthetic embeddings}. Figure~\ref{fig:embd} presents a 2D UMAP projection comparing embeddings of Clotho from real data with synthetic data from three different models: a baseline Random approach, the DM method, and ImageBindDC. The Random method on the left fails to capture the diversity of the real data, as its synthetic embeddings are confined to a small, concentrated cluster. The DM method in the center demonstrates better spread, yet its synthetic embeddings still form distinct clusters that are largely separate from the real data clouds. In contrast, ImageBindDC proposed method on the right generates synthetic audio and text embeddings that are well-integrated and distributed throughout the corresponding real embedding spaces, indicating a much closer match to the true data distribution and showcasing the superiority of our approach.

\section{Conclusion}
In this work, we introduced ImageBindDC, a novel framework that significantly advances multi-modal data condensation. Our approach overcomes the limitations of existing methods by operating in a unified feature space and employing a powerful Characteristic Function Discrepancy for more precise distribution matching. We designed a multi-faceted objective to enforce uni-modal, cross-modal, and joint-modal consistency, ensuring that the intricate relationships between modalities are preserved. Experiments confirm the effectiveness of ImageBindDC, demonstrating its ability to synthesize high-quality data.

\bibliography{colm2024_conference}
\bibliographystyle{colm2024_conference}

\clearpage

\appendix

\section{Detailed Experiment settings}
\label{app:details}

\subsection{Baselines} 

We conducted experiments on both data selection and distillation methods. The selection-based methods we utilized are as follows:
\begin{itemize}
    \item \textbf{Random}: A baseline method that selects data points randomly.
    \item \textbf{Herding}~\citep{welling2009herding}: This method uses a greedy approach to select data points that minimize the discrepancy between the gradients of the data.
    \item \textbf{Forgetting}~\citep{toneva2018empirical}: This method gradually removes the least informative samples and selects the most important ones.
    \item \textbf{GraNd}~\citep{paul2021deep}: Optimizes a synthetic dataset by minimizing the divergence between the gradients of the model trained on the synthetic dataset and those trained on the original data.
\end{itemize}
For data distillation methods, we considered the following:
\begin{itemize}
    \item \textbf{DC}~\citep{DC}: This method synthesizes a compact dataset that retains the essential information of the original data.
    \item \textbf{DSA}~\citep{DSA}: An enhancement to DC, DSA incorporates differentiable augmentation to make better use of data transformations.
    \item \textbf{MTT}~\citep{MTT}: This method aligns the training trajectories of models trained on synthetic and real data, ensuring that both models follow similar optimization paths.
    \item \textbf{DM}~\citep{DM}: Aligns the distributions of real and synthetic data in the feature space. While this provides computational efficiency, it may miss higher-order moments of the data distribution, leading to less precise condensations.
    \item \textbf{AVDD}~\citep{kushwaha2024audio}: Specifically designed for audio-visual data condensation, this method aims to optimize the synthesis process for such multimodal data.
\end{itemize}

\subsection{Model Training}

For the experiment using the audio-visual modality, we trained the model with the following parameters. When computing CFD, we adopted Gaussian distribution for $t$. The number of evaluation steps was set to 3. Image and audio augmentation were applied by enabling the image-domain modulation augmentation. The synthetic data initialization was performed using the herding method. The model was trained for 30 iterations, with evaluations conducted every 10 steps.

The learning rates for updating the synthetic audio and synthetic images were both set to 0.5. The classifier's learning rate was set to 0.001, while the learning rates for the frame and sound modalities were set to 0.0001 and 0.001, respectively. The Adam optimizer was used with a beta1 value of 0.9, and a weight decay of 0.0001 was applied to the parameters.

We used a batch size of 32 for synthetic data and 128 for real data. Additionally, four data loader workers were used to optimize the data loading process. For data augmentation, a differentiable Siamese augmentation strategy was employed, which included techniques such as color adjustments, cropping, cutout, flipping, scaling, and rotation.

In terms of model architecture, the sound modality was processed using a convolutional network, and similarly, the frame modality was processed using a convolutional network as well. The classifier was implemented using an ensemble architecture.

For loss functions, the base distribution matching loss was assigned a weight of 1.0, and the loss function parameters were both set to 0.5. The model was trained for 30 epochs, and evaluations were performed at intervals of 1000 steps. The results were stored for further analysis after each evaluation.

\subsection{Metrics}

We use Recall\@K to evaluate the audio-text retrieval performance. Recall\@K is defined as the fraction of relevant items retrieved within the top-K results for a given query.

For a query $q$ and a set of $N$ candidate items, Recall\@K is computed as:
\begin{equation}
\text{Recall\@K} = \frac{\text{Number of relevant items in top-K}}{\text{Total number of relevant items}}
\end{equation}
For audio-to-text retrieval (A2T) and text-to-audio retrieval (T2A), the recall at rank $K$ is given by:
\begin{equation}
R_{\text{A2T}}@K = \frac{\sum\_{i=1}^{N} \mathbb{I}(q\_{\text{audio}}, t\_i \in \text{top-K})}{N}
\end{equation}
\begin{equation}
R_{\text{T2A}}@K = \frac{\sum\_{i=1}^{N} \mathbb{I}(q\_{\text{text}}, a\_i \in \text{top-K})}{N}
\end{equation}
where $\mathbb{I}(\cdot)$ is an indicator function, which equals 1 if the relevant item is within the top-K retrieved results, and 0 otherwise.

\subsection{Details of Evaluation Computational Efficiency}
We evaluated the computational efficiency of our method on the VGGS-10K dataset. The evaluation included peak GPU memory usage (GPU Memory in GB) and training time (Condensation Time) at different DPC values of 1, 10, and 20. All experiments were conducted on a single A100 GPU with 80GB of memory.

\subsection{Encoding Architecure}
For the ConvNet architecture, we use a standard approach \cite{ConvNet}, where the audio input is processed through 3 blocks, each consisting of convolution, normalization, ReLU, and pooling layers. For visual inputs we use 5 such blocks. We conduct experiments on both architectures to evaluate their performance in the context of multi-modal dataset condensation. Both architectures were trained using pre-trained parameters, with only the final linear layer being fine-tuned.

\begin{algorithm*}[tb!]
\caption{Pseudo code for our proposed pipeline ImageBindDC. We take image and audio data as an illustration.}
\label{alg:imagebinddc}
\begin{algorithmic}[1]
\STATE \textbf{Input:} Real multi-modal dataset $\mathcal{R} = \{(x_a, x_v, y)\}_{i=1}^N$, where $x_a$ is audio data, $x_v$ is image data, and $y$ are the corresponding labels
\STATE \textbf{Output:} Synthetic multi-modal dataset $\mathcal{S} = \{(\tilde{x}_a, \tilde{x}_v, y)\}_{j=1}^M$, where $M \ll N$
\STATE Initialize: Pre-trained ImageBind encoder $\psi$ for embedding both audio and image modalities
\STATE \textbf{Repeat for each training iteration:}
\FOR{each batch in $\mathcal{R}$}
    \STATE Extract real audio and image data: $(x_a, x_v) \in \mathcal{R}$
    \STATE Obtain embeddings: $e_a = \psi_a(x_a)$, $e_v = \psi_v(x_v)$
    \STATE Sample synthetic data $(\tilde{x}_a, \tilde{x}_v) \in \mathcal{S}$
    \STATE Obtain synthetic embeddings: $\tilde{e}_a = \psi_a(\tilde{x}_a)$, $\tilde{e}_v = \psi_v(\tilde{x}_v)$
    \STATE Compute losses:
        \STATE \quad \textit{Uni-modal Alignment:} $\mathcal{L}_{\mathrm{audio}} = \mathrm{CFD}(e_a, \tilde{e}_a)$, $\mathcal{L}_{\mathrm{image}} = \mathrm{CFD}(e_v, \tilde{e}_v)$
        \STATE \quad \textit{Cross-modal Alignment:} $\mathcal{L}_{\mathrm{cross}} = 1 - \rho_\mathrm{cross}$, where $\rho_\mathrm{cross}$ is the similarity between real and synthetic data
        \STATE \quad \textit{Joint-Modal Alignment:} $\mathcal{L}_{\mathrm{joint}} = 1 - \rho_\mathrm{joint}$, where $\rho_\mathrm{joint}$ is the matrix similarity of mean embeddings

    \STATE Compute total loss: 
    \[
    \mathcal{L} = \lambda_{\mathrm{uni}} \mathcal{L}_{\mathrm{uni}} + \lambda_{\mathrm{cross}} \mathcal{L}_{\mathrm{cross}} + \lambda_{\mathrm{joint}} \mathcal{L}_{\mathrm{joint}}
    \]
    \STATE Update synthetic dataset $\mathcal{S}$ using gradient descent on the total loss
\ENDFOR

\STATE \textbf{Return:} The condensed synthetic multi-modal dataset $\mathcal{S}$
\end{algorithmic}
\end{algorithm*}

\subsection{Datasets}

\begin{itemize}
    \item \textbf{AVE}~\cite{tian2018audio}: We segmented each clip into non-overlapping one-second windows aligned with synchronized annotations, resulting in train/val/test splits of 27,726, 3,288, and 3,305 samples, respectively. 
    \item \textbf{VGGS-10K}~\citep{VGG}: The dataset is derived from VGGSound~\citep{chen2020VGGSound}. For experiments, we selected the central one-second video from each original clip in the train/test splits, resulting in approximately 165k and 13k samples, respectively. Then we randomly selected a subset of 10 classes, referred to as VGGS-10k, which contains 8,808 training videos and 444 test videos.
    \item \textbf{NYU-v2}~\citep{NYU}: This is a large-scale multi-modal dataset consisting of 1,449 samples, with 795 samples designated for training and 654 samples for testing.
    \item \textbf{Clotho}~\citep{Clotho}: The dataset consists of 2,893 audio clips in the development set, which are used for training and distillation. For experiments, we randomly split the dataset into training and validation sets with an 80/20 ratio, resulting in 2,314 samples for training and 579 samples for validation.
\end{itemize}

\subsection{Pseudocode for ImageBindDC}

Algorithm \ref{alg:imagebinddc} details the procedure for synthesizing a condensed dataset with a specified number of Datapoints Per Class (DPC).

\end{document}